\def\year{2019}\relax
\documentclass[letterpaper]{article} %DO NOT CHANGE THIS
\usepackage{aaai19}  %Required
\usepackage{times}  %Required
\usepackage{helvet}  %Required
\usepackage{courier}  %Required
\usepackage{url}  %Required
\usepackage{graphicx}  %Required
\usepackage{multirow}
\usepackage{amsmath}
\usepackage{caption}
\usepackage{booktabs}
\usepackage{subcaption}
\usepackage{hhline}
\newcommand{\citet}[1]
{\citeauthor{#1}~\shortcite{#1}}
\newcommand{\citep}{\cite}

\begin{document}
\title{Unsupervised Transfer Learning for Spoken \\Language Understanding in Intelligent Agents}
\author{Aditya Siddhant \textsuperscript{1}, Anuj Kumar Goyal \textsuperscript{2}, Angeliki Metallinou \textsuperscript{2}\\
asiddhan@cs.cmu.edu, anujgoya@amazon.com, ametalli@amazon.com\\
\textsuperscript{1} Carnegie Mellon University, \textsuperscript{2} Amazon Alexa AI
}
\maketitle
\begin{abstract}
User interaction with voice-powered agents generates large amounts of unlabeled utterances. In this paper, we explore techniques to efficiently transfer the knowledge from these unlabeled utterances to improve model performance on Spoken Language Understanding (SLU) tasks. We use Embeddings from Language Model (ELMo) to take advantage of unlabeled data by learning contextualized word representations. Additionally, we propose  ELMo-Light (ELMoL), a faster and simpler unsupervised pre-training method for SLU. Our findings suggest unsupervised pre-training on a large corpora of unlabeled utterances leads to significantly better SLU performance compared to training from scratch and it can even outperform conventional supervised transfer. Additionally, we show that the gains from unsupervised transfer techniques can be further improved by supervised transfer. The improvements are more pronounced in low resource settings and when using only 1000 labeled in-domain samples, our techniques match the performance of training from scratch on 10-15x more labeled in-domain data.
\end{abstract}

\section{Introduction}
\label{sec:introduction}

Voice-powered artificial virtual agents have become popular amongst consumer devices, as they enable their users to perform everyday tasks through intuitive and natural user interfaces. SLU tasks such as intent classification and entity tagging are critical functionalities of these agents. Fast expansion of these functionalities to new domains is important for achieving engaging and informative interactions, as it increases the range of capabilities that their users enjoy.

For SLU tasks, most of the current methods use supervised learning, which relies on manually labeled data for building high quality models. The supervised learning paradigm is therefore costly, time-consuming and does not scale well for cases where the label space is continuously expanding as new functionality is added to an agent. Also, user interaction with voice-powered agents generates large amounts of unlabeled text, produced by the Automatic Speech Recognition (ASR) engine. This ASR output text is a large and valuable resource of conversational data that is available in practically unlimited quantities and could be used to improve the agent's SLU accuracy. Thus, the ability to learn effectively from unlabeled text is crucial to alleviating the bottlenecks of supervised learning.

The machine learning community is actively exploring transfer learning and unsupervised learning for low resource tasks. \citet{goyal_tl_2018} explored transfer learning from existing annotated SLU domains for building models for related, low-resource domains for artificial agents. However, such transfer learning techniques rely on large annotated resources from a related functionality. Recent work has used language modeling (LM) as a proxy task for learning context dependent word embeddings from large unlabeled text corpora \cite{peters_elmo_2018}. These embeddings allow for unsupervised knowledge transfer and have been shown to bring performance gains for various downstream natural language processing (NLP) tasks. 

In this work, we propose an unsupervised transfer learning technique inspired from ELMo and Universal Language Model Fine Tuning (ULMFiT) to leverage unlabeled text for building SLU models \cite{peters_elmo_2018,howard_ruder_2018}. We also explore the combination of unsupervised and supervised knowledge transfer for SLU. We evaluate our methods on various tasks and datasets, including data from a popular commercial intelligent agent. Our results show that unsupervised transfer using unlabeled utterances can outperform both training from scratch and supervised pre-training. Additionally, the gains from unsupervised transfer can further be improved by supervised transfer. These improvements are more pronounced in low resource setting and when only 1K labeled in-domain samples are available, the proposed techniques match the performance of training from scratch on 10-15x more labeled data. Concretely, our contributions are:
\begin{itemize}
\item We apply ELMo embeddings for unsupervised knowledge transfer from raw ASR text and show SLU accuracy gains.
\item We propose ELMo-Light (ELMoL), a light-weight ELMo alternative that is well-suited for commercial settings, with comparable accuracy to ELMo for most SLU tasks.
\item We combine unsupervised and supervised transfer learning, and show the additive effect of the two techniques.
\item We extensively evaluate our methods on benchmark SLU datasets and data from a commercial agent, across various resource conditions.
\end{itemize}

The rest of paper is organized as follows. To provide a bit of background, we discuss related work and neural architectures for SLU and then introduce the methods we use for unsupervised transfer including the proposed ELMoL. Finally, we describe the datasets, experimental setup, results and end with directions for future work. Table \ref{tab:notation} summarizes some of the frequently used abbreviations throughout the paper.

\begin{table}[h] 
\centering
\begin{tabular}{|c|c|}
\hline
Abbr. & Description                                                                                      \\ \hline
UT    & Unsupervised Transfer                                                                            \\ \hline
ST    & Supervised Transfer                                                                              \\ \hline
IC    & Intent Classification                                                                            \\ \hline
ET    & Entity Tagging                                                                                   \\ \hline
LM    & Language Model                                                                                   \\ \hline
\textit{guf}   & Gradual Unfreezing                                                                               \\ \hline
\textit{tlr}   & Triangular Learning Rate                                                                         \\ \hline
\textit{discr} & Discriminative Fine Tuning                                                                       \\ \hline
\end{tabular}
\caption{Abbreviation Table} \label{tab:notation}
\end{table}

\section{Related Work}
\label{sec:related}
 
Deep learning models using CNNs and LSTMs are state of the art for many NLP tasks. Examples include applying LSTMs for sentence classification \cite{liu_multi_timescale,Socher_recursive_2013}, LSTM with Conditional Random Field (CRF) decoder for sequence labeling \cite{chiu2016named} and CNN-LSTM combinations for LM \cite{jozefowicz2016exploring}. LSTMs with attention have also been used for SLU tasks including Entity tagging (ET) and intent classification (IC) \cite{liu_lane2016}. 

To enable robust training of deep learning models in low resource settings, the community is actively exploring semi-supervised, transfer and multi-task learning techniques. In the multi-task paradigm a network is jointly trained to optimize multiple related tasks, exploiting beneficial correlations across tasks \cite{liu_lane2016,collobert_weston_08}. \citet{liu_lm_2018} used language models (LMs) as an auxiliary task in a multi-task setting to improve sequence labeling performance. Transfer learning addresses the transfer of knowledge from data-rich source tasks to under-resourced target tasks. Neural transfer learning has been successfully applied in computer vision where lower network layers are trained in high-resource supervised datasets like ImageNet to learn generic features \cite{krizhevsky2012imagenet}, and are then fine-tuned on target tasks, leading to impressive results for image classification and object detection \cite{donahue2014decaf,sharif2014cnn}. In NLP, such supervised transfer learning was successfully applied for SLU tasks, by learning IC and ET models on high resource SLU domains, and then fine-tuning the network on under resourced domains \cite{goyal_tl_2018}. Similar ideas have also been explored for POS tagging using for cross-lingual transfer learning \cite{kim_emnlp_2017}.

Unsupervised methods for knowledge transfer include computing word and phrase representations from large unlabeled text corpora. Examples include Word2Vec and FastText, where context independent word representations are learnt based on LM-related objectives \cite{mikolov_w2v_2013,bojanowski2017enriching}. Unsupervised sentence representations have been computed via predicting sentence sequences like skip-thought \cite{kiros_skipthought_2015}, and through a combination of auxiliary supervised and unsupervised tasks \cite{Cer2018universal}. Recent work has introduced LM-based word embeddings, ELMo, that are dependent on sentence context and are shown to lead to significant accuracy gains for various downstream NLP tasks \cite{peters_elmo_2018}. Unsupervised pre-training has also been used as a form of knowledge transfer by first training a network using an LM objective and then fine-tuning it on supervised NLP tasks. This has been shown to be efficient for sentence classification \cite{howard_ruder_2018,dai_le_2015} and \cite{Radford_2018} for textual entailment and question answering. Our work, building upon transfer learning ideas such as supervised model pre-training \cite{goyal_tl_2018}, LM-fine tuning \cite{howard_ruder_2018} and context dependent word embeddings \cite{peters_elmo_2018}, introduces a light-weight ELMo extension and combines those methods for improving SLU performance in a commercial agent.

\section{Neural Architectures for SLU} 
\label{sec:neural_slu}

We focus on SLU for voice powered artificial agents, specifically on intent classification (IC) and Entity tagging (ET) models which are essential for such agents. Given a user request like `how to make kadhai chicken', the IC model classifies the intention of the user, such as `GetRecipe' while the ET model tags the entities of interest in the utterance, such as `Dish'=`kadhai chicken'.

\begin{figure}[t]
  \includegraphics[width=\linewidth]{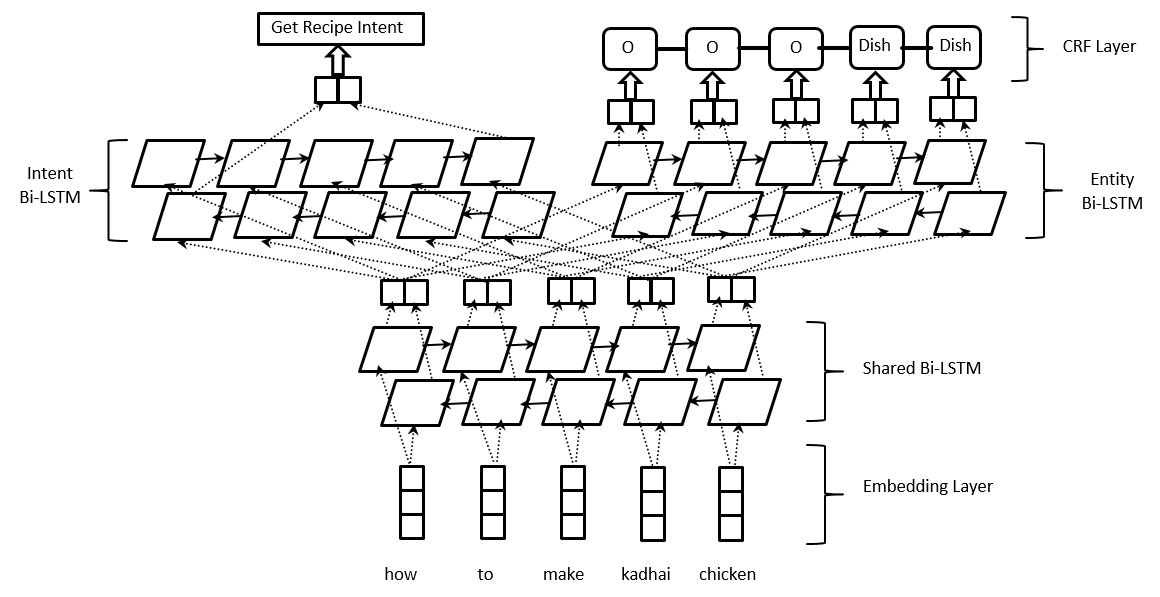}
  \caption{Multi-task architecture for IC and ET}
  \label{fig:architecture_supervised_TL}
\end{figure}

We use a multi-task deep neural network architecture for jointly learning the IC and ET models, hence exploring beneficial correlations between the two tasks. Our architecture is illustrated in Figure \ref{fig:architecture_supervised_TL}. It consists of a bottom shared bidirectional LSTM (bi-LSTM) layer on top of which we train a bi-LSTM-CRF for ET and a bi-LSTM for IC. The two top layers are optimized separately for ET and IC, while the common bottom layer is optimized for both tasks. The objective function for the multi-task network combines the IC and ET objectives. 

Specifically let $r^{c}_t$ denote the common representation computed by the bottom-most bi-LSTM for each word input at word $t$. The ET forward LSTM layer learns a representation $r^{ET, f}_t = \phi ( r^c_t, r^{ET}_{t-1})$, where $\phi$ denotes the LSTM operation. The IC forward LSTM layer learns $r^{IC, f}_t = \phi ( r^c_t, r^{IC}_{t-1})$. Similarly, the backward LSTM layers learn $r^{ET, b}_t$ and $r^{IC, b}_t$. 

To obtain the entity tagging decision, we feed the ET bi-LSTM layer's output per step, denoted as $[r^{entity}_t]_{t=1}^T$ into the CRF layer, and produce a entity label sequence $[\hat{S}_t]_{t=1}^T$. For the intent decision, we concatenate the last step from the forward LSTM with the first step of the backward LSTM to get the intent representation $r^{intent}$, and feed it into a softmax layer for classification:

\begin{gather*}
r^{entity}_t = r^{ET,f}_{t} \oplus r^{ET,b}_{t}, r^{intent} = r^{IC,f}_{T} \oplus r^{IC,b}_{1} \\
[ \hat{S}_t ]_{t=1}^T=CRF( [ r^{entity}_t]_{t=1}^T ) \\
\hat{I}=softmax( W_I r^{intent} + b_I)
\end{gather*}
where $\oplus$ denotes concatenation. $W_I, b_I$ are the weights and biases for the intent softmax layer and $CRF(\cdot)$ denotes the CRF layer. $\hat{S}_t$ is the predicted entity tag per step, and $\hat{I}$ is the predicted intent label for the utterance.

\section{Methods for Unsupervised Transfer}
\label{sec:methods}
When building ET and IC models for a target new SLU domain like recipes, we typically have between few hundreds and few hundred thousand labeled utterances available for supervised training. We also have manually labeled ET and IC data from existing source SLU domains, like music, weather etc, which is a few million combined. Finally, we have billions of unlabeled utterances produced by the ASR engine from live user-agent interaction in our system. In this section, we describe Unsupervised Transfer (UT) learning techniques that leverage this large corpus of unlabeled ASR text to improve the accuracy of the IC and ET tasks. We also combine UT with supervised transfer learning from the existing labeled data from source domains, to obtain additional gains. 

\subsection{Embeddings from Language Model (ELMo)}
\label{sec:elmo}

\citet{peters_elmo_2018} introduced contextualized word embeddings, called ELMo, computed by first training an LM using a state-of-the-art architecture with multiple CNNs over characters and $L$ bi-LSTM layers on top to model the sentence (the network is also referred to as CNN-BIG-LSTM \citep{jozefowicz2016exploring}). After the model is trained on a large corpus, the outputs of the different layers are used as embeddings for downstream tasks. Specifically, let's denote with $x_t$ the context-independent word representations from the CNN layer at word $t$ and with $h_{t,i}^{LM}$ the contextual word representations obtained by concatenating the forward and backward LSTM activations at word $t$ and layer $i$. The contextual ELMo embeddings at $t$ are computed through linear combination as:
\begin{gather*}
ELMo_t=\gamma ( s_0 \cdot x_t + \Sigma_{i=1}^L s_i \cdot h_{t,i}^{LM})
\end{gather*}
where $\gamma$ is a scalar parameter for scaling the ELMo vector

We trained ELMo embeddings using $L=2$ bi-LSTM layers on ASR text, and used them as input to the downstream SLU architecture of Figure \ref{fig:architecture_supervised_TL}. ELMo parameters were kept frozen, as in \citet{peters_elmo_2018}, and parameters $s_i$ and $\gamma$ were jointly optimized on the ET and IC tasks. An advantage of this unsupervised pre-training is that the CNN-BIG-LSTM weights do not experience catastrophic forgetting, therefore the SLU architecture can be trained without losing the knowledge gained from unlabeled data. However, computing ELMo embeddings at runtime introduces many additional parameters (e.g., weights from the large CNN-BIG-LSTM). In practice, we observed that using ELMo increases the inference time by 1.6x and triples the memory requirement at runtime because of additional parameters, which makes this solution less attractive for a commercial SLU system.

\begin{figure*}[h]
       \centering
        \begin{subfigure}[b]{0.45\textwidth}
                \centering
                \includegraphics[width=\linewidth]{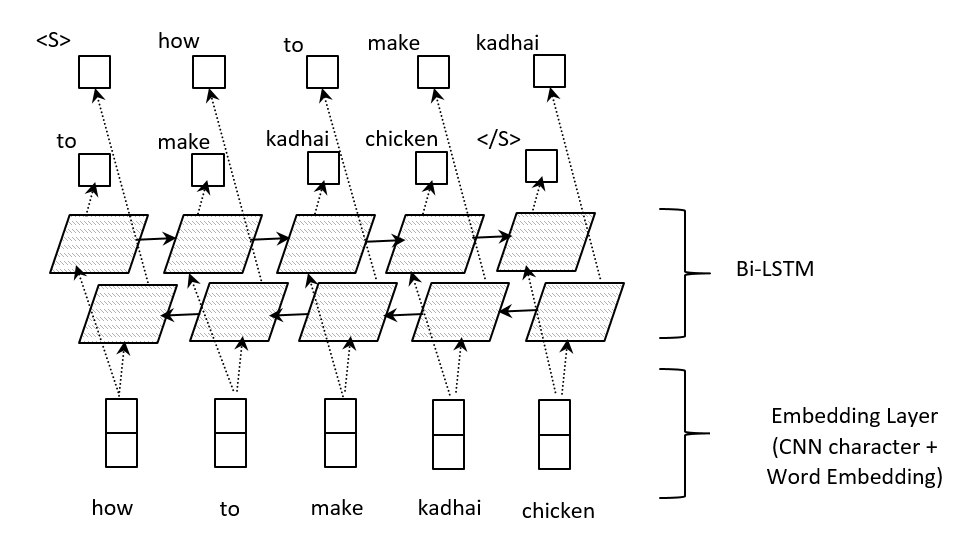}
                \caption{ELMoL LM Pre-training}
                \label{fig:sllmo_lm_pretrain}
        \end{subfigure}%
        \begin{subfigure}[b]{0.5\textwidth}
                \centering
                \includegraphics[width=\linewidth]{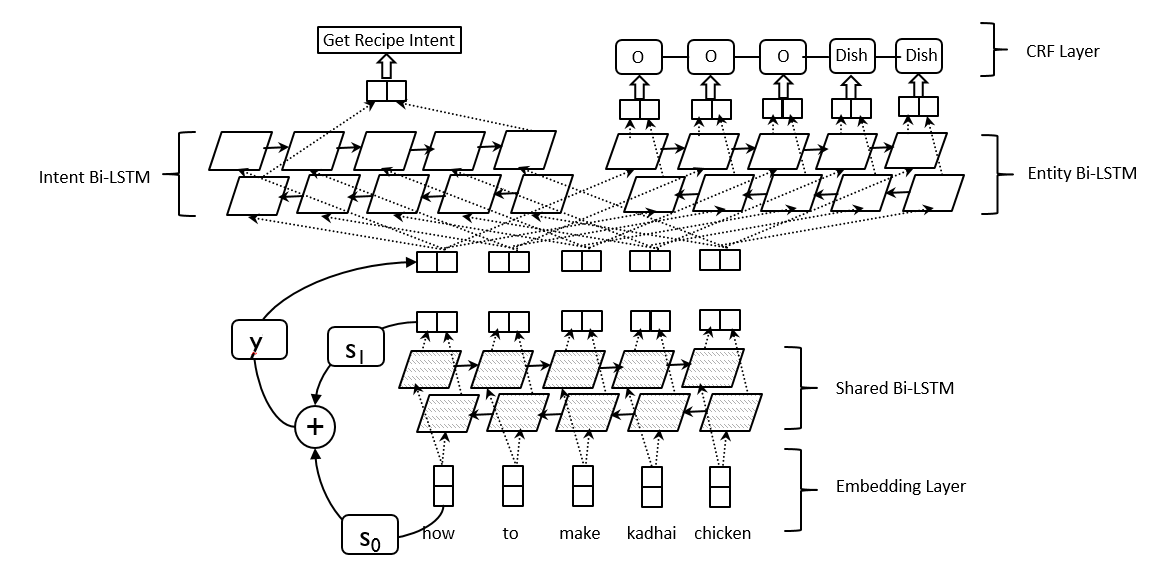}
                \caption{ELMoL embeddings used in the SLU multi-task model}
                \label{fig:sllmo_slu}
        \end{subfigure}%
        \caption{Left: ELMoL embeddings from simplified LM. Right: ELMoL embeddings used in the SLU model}
        \label{fig:sllmo_embeddings}
\end{figure*}

\subsection{ELMo-Light (ELMoL) for SLU tasks}
\label{sec:sslmo}

We introduce ELMo-Light (ELMoL) which is a light-weight alternative for computing contextualized word embeddings within our SLU architecture. We base our model on two observations. First, typical conversational requests to artificial agents use brief and simple language therefore a smaller LM architecture may be sufficient. Second, the purpose of the lower shared layer of our SLU is to learn generic representations for IC and ET, and could be additionally pre-trained with an LM objective on a larger corpus. 

Specifically, we train an LM network consisting of character level CNNs and a single bi-LSTM layer on unlabeled ASR text to compute contextual word representation $h_{t,i}^{LM}$ at word $t$, as shown in Fig \ref{fig:sllmo_embeddings}(a). Then, we combine these contextual embeddings with non-contextual word embeddings $x_t$, and feed them as input to our upper SLU model layers for IC and ET, optimizing the linear weights on the downstream tasks, similarly to ELMo:
\begin{gather*}
ELMoL_t=\gamma ( s_0 \cdot x_t + s_1 \cdot h_{1,i}^{LM})
\end{gather*}

This model is illustrated in Fig \ref{fig:sllmo_embeddings}(b). Compared to our original architecture of Fig \ref{fig:architecture_supervised_TL}(a), the proposed ELMoL model introduces only three additional trainable parameters ($\gamma, s_0, s_1$). This is because the single bi-LSTM LM weights are re-purposed as the lower layer our of SLU architecture and are further fine-tuned on the in-domain data along with the rest of the SLU network. The character level CNNs of the LM are removed from the final architecture and their output is used as fixed non-trainable embeddings. Therefore, ELMoL embeddings do not incur additional latency or memory costs and are well suited for our SLU system.

\subsubsection{Techniques for effectively training ELMoL}

In contrast to ELMo, for the ELMoL model we do not freeze the weights of lower bi-LSTM layer. Freezing them would only leave a single trainable bi-LSTM layer for each of the IC and ST tasks during the supervised training stage, which empirically negatively impacts the final accuracy. Instead, after unsupervised LM pre-training is complete for the lower layers, we update all layers of the network using the target domain labeled IC and ET data. 

We empirically noticed that making the lower layer trainable causes catastrophic forgetting, e.g., the knowledge transferred through pre-trained weights is erased by the gradient updates during supervised training. To counter this issue, we apply a combination of techniques from the literature \citep{howard_ruder_2018}, specifically \textbf{gradual unfreezing (\textit{guf})}, \textbf{discriminative fine-tuning (\textit{discr})} and \textbf{slanted triangular learning rates (\textit{tlr})}. \textit{guf} means updating only the top layer for a few epochs keeping lower layers frozen, and then progressively updating bottom layers. It can be combined with \textit{discr}, e.g, using different learning rates across network layers. Here, we use a lower learning rate for the bottom layer to avoid large updates in the transferred knowledge of the lower layer weights. \textit{tlr} refer to learning rates that are initially slow which discourages drastic updates in early stages of learning, then it rapidly increase allowing more exploration of the parameter space and then slowly decrease enabling learned parameters to stabilize.

%rapidly increase to allow learning to explore various areas of the parameter space, and then slowly decrease to enable stabilizing the learned parameters.

\subsection{Combining Supervised Transfer (ST) with UT}
\label{sec:combined}

\citet{goyal_tl_2018} used supervised transfer learning in a similar SLU model by pre-training a multi-task network on labeled data from source domains and fine-tuning the weights on ET and IC tasks of an under-resourced target domain, updating the top layers to reflect the labels of the new domain. We combine this supervised pre-training technique with unsupervised ELMo and our proposed ELMoL embeddings. We explore the following two variations for combining unsupervised pre-training with supervised transfer learning.

\subsubsection{ELMo+ST} This method is similar to \citet{goyal_tl_2018}, the only difference being that instead of FastText embeddings, we use ELMo embeddings while pre-training the network on labeled data from source domain(s). This method, therefore, becomes a 3 step process. The first step involves training the external LM network using unlabeled data. In the second step, we use the embeddings from LM, trained in the first step, to pre-train our multi-task IC/ET architecture using labeled data from source domains. Finally, in the third step, we fine tune the weights in multi-task IC/ET architecture on labeled data from target domain. The LM weights are frozen in steps 2 and 3.

\subsubsection{ELMoL+ST}
This is also a three step method where in the first step, the lower layer of multi-task IC/ET architecture is pre-trained using a language model objective. This step is essentially the same as the first step of ELMoL training without ST. In the second step, we train the entire architecture on labeled source data using \textit{guf} and \textit{discr}. This step, therefore, becomes a two step process itself in which we first train the upper layers of the architecture keeping the lower layer frozen with weights from step 1. Once the upper layers stabilize, we train the entire network with labeled source data, keeping the lower layer learning rate much lower than the higher layer (\textit{discr}). Finally, in the third step of the process, we fine tune all the layers on labeled data from target domain using all 3 techniques \textit{guf}, \textit{discr} and \textit{tlr}. 

While we combine ST with UT to see if it can provide an additive effect and bump performance even further, we would like to point out that the main focus of this paper is unsupervised transfer learning and therefore, comparison of ST+UT with other semi-supervised techniques do not form a part of our experiments.

\section{Datasets}
\label{sec:datasets}

\subsection{Internal Labeled and Unlabeled SLU datasets}
\label{sec:internal_datasets}

We use two internal SLU domains as our target domains, denoted as \textbf{Domain A} (5 intents, 36 entities) and \textbf{Domain B} (22 intents, 43 entities). We have a total of 43K and 100K labeled training samples for Domains A and B respectively. In addition to these two target labeled datasets, we pool labeled datasets from tens of other domains of our SLU system. In total, we use around 4 million utterances which are labeled in terms of the intents and entities of their respective domains. We refer to this labeled dataset as \textbf{Internal Labeled Source Dataset (ILSD)}. ILSD spans data from a range of functionalities like playing media (music, movies, books), question answering, performing tasks like setting calendar events, alarms and notifications, etc. The total number of intents and entities in this dataset are of the order of hundreds each.  We also collect unlabeled ASR text from the live utterances of users interacting with our agent. Specifically, we use a large set of 250 million de-duplicated tokens collected over a year. We refer to this dataset as \textbf{Internal Unlabeled Dataset (IUD)}.

\subsection{Public Labeled and Unlabeled datasets}
\label{sec:benchmark_SLU}

For benchmarking our proposed methods we use two labeled public SLU datasets: \textbf{ATIS} and \textbf{SNIPS}. \textbf{ATIS} is a common SLU benchmark from the travel planning domain which contains 5K utterances \cite{hemphill1990atis}. \textbf{SNIPS} is a more recent SLU benchmark created by the company \textit{snips.ai} for benchmarking commercial NLU engines offered by companies like Google, Amazon, etc \cite{snips_url}. \textbf{SNIPS} includes 7 commonly used intents like asking for the weather, playing music, booking restaurants and contains 13K training utterances. We also used two unlabeled datasets, the \textbf{1B Word Benchmark (1BWB)} and the \textbf{1M SLU Benchmark data (1MSLU)}. The 1BWB is a common text benchmark dataset used in the LM literature \cite{chelba2014_1BWB}, it is extracted from News Crawl and contains 750 million tokens. We additionally collected an SLU text dataset by combining various public SLU corpora, including training splits from ATIS, SNIPS, DSTC2 \citep{henderson2014second}, and others. We call this dataset 1M SLU Benchmark data (1MSLU) and it contains 1 million tokens in total. These datasets were used as additional resources to the internal ASR data for training language models for unsupervised knowledge transfer. Statistics of the target datasets are in Table \ref{table:data_stat}.

\begin{table}[h]
\scalebox{0.75}{
\begin{tabular}{|c|c|c|c|c|c|c|}
\hline
         & \begin{tabular}[c]{@{}c@{}}\#Training \\ Samples\end{tabular} & \begin{tabular}[c]{@{}c@{}}\#Dev \\ Samples\end{tabular} & \begin{tabular}[c]{@{}c@{}}\#Test\\ Samples\end{tabular} & \begin{tabular}[c]{@{}c@{}}Vocab\\ Size\end{tabular} & \#Int. & \#Ent. \\ \hline
Domain-A & 43168                                                         & 3680                                                    & 4752                                                     & 62600                                                & 5         & 36         \\ \hline
Domain-B & 100000                                                        & 8227                                                    & 8695                                                     & 62600                                                & 22        & 43         \\ \hline
ATIS     & 4478                                                          & 500                                                     & 893                                                      & 868                                                  & 26        & 79         \\ \hline
SNIPS    & 13084                                                         & 700                                                     & 700                                                      & 11823                                                & 7         & 39         \\ \hline
\end{tabular}}
\caption{Internal and public target datasets statistics} \label{table:data_stat}
\end{table}

\section{Experiments and Results}
\label{sec:experiments_results}
%We introduce the baselines, lay out the experimental setup and present our results.

\subsection{Baselines}
\label{sec:experiment_baselines}
Our first baseline uses no form of unsupervised pre-training and works only with the target data at hand. We refer to this baseline as \textbf{NoUT}. Our second baseline uses pre-trained FastText word embeddings \cite{bojanowski2017enriching}. We chose this baseline because pre-trained word embeddings is a very popular means of unsupervised pre-training in the NLP literature. This baseline is referenced as \textbf{Fasttext}.

\subsection{Experimental Setup}
\label{sec:experiment_setup}

\subsubsection{Network Hyper-parameters}
The experimental setup was kept same for all 4 datasets with minor adjustments in hyperparameters. We use 200 hidden units for all three LSTM layers in our multi-task architecture. We use Adam optimizer \cite{kingma2015method} with initial learning rate 0.0001 for internal datasets (Domain-A and Domain-B) and 0.0005 for ATIS and SNIPS. Both IC and ET losses are weighted equally in the total loss. Training is done upto 25 epochs with early-stopping based on sum of IC and ET scores on development set. Dropout probability is 0.5 for ATIS and SNIPS and 0.2 for internal datasets and we use L2 regularization on all weights with lambda=0.0001. 

\subsubsection{Network Input Embeddings}
The embedding layer dimension depends on the type of embedding being used. For NoUT and FastText, the embedding dimension is 400. NoUT has all 400 trainable while FastText has 300 pre-trained and 100 trainable. ELMo uses 1024 dimension embedding while ELMoL has 200 dimension embedding. ELMoL embedding is reduced to 200 in order to keep the number of parameters same as NoUT or FastText. This is because to linearly combine embedding and LSTM output, the embedding dimension has to be the same as the forward and backward LSTM output. First 100 dimensions of those embeddings are word-level trainable embeddings and the rest are fixed output of character-level CNN.

\subsubsection{LM Training for ELMo}
To train the LM for ELMo, we follow \citet{peters_elmo_2018} very closely. For internal datasets, we train a CNN-BIG-LSTM LM, halving the embedding and hidden layer dimensions, on IUD from scratch for 10 epochs. As an alternative, we also fine-tune the LM already trained on 1B Word Benchmark on IUD for 5 epochs. The LM to be used for embeddings is decided based on perplexity on a heldout set (1\% of IUD). For ATIS and SNIPS, we use 1M SLU Benchmark to both train from scratch and fine-tune the LM trained on 1B Word Benchmark for 25 and 10 epochs respectively.

\subsubsection{LM Training for ELMoL}
ELMoL language model training is carried out similar to ELMo. The main difference is in the LM architecture being used. Keeping it similar to the lower layer of multi-task model, there is only 1 LSTM layer with 200 hidden units. Context independent CNN layer has 10, 20, 20, 20, 20 and 10 channels of width 1,2,3,4,5 and 6 respectively. For internal domains, the model was trained on IUD with a batch size of 128 for 25 epochs. For ATIS and SNIPS, we train the on 1M SLU Benchmark with batch size of 32 for 50 epochs. While using \textit{guf}, there are two important hyper-parameters, learning rate for first round of training (when lower layer is frozen) which we keep as 0.0001 for internal datasets and 0.0005 for public datasets. The second important hyper-parameter is the number of epochs after which lower layer is unfrozen. This is usually between 10-15 epochs. The learning rate for second round of training (with all layers unfrozen) is decided based on the dev performance and varies across datasets. We also use \textit{discr} and \textit{tlr} in second round of training. For \textit{discr}, we use the setting followed by \citet{howard_ruder_2018} and keep the lower layer learning rate 2.5 times lesser than the upper layer. For \textit{tlr}, we keep the starting learning rate 10 times lesser than the peak learning rate and this peak is located at 1/8th of the total number of updates. 

\subsubsection{Simulating Low Resource Data Settings}
We simulate low resource settings by carrying out experiments on smaller training sets sampled from all four target datasets (Domain A, Domain B, ATIS, SNIPS). Samples of size 100, 200, 500, 1000, 2000, 5000 and 10000 are drawn from the target training sets. We then compare the different methods using smaller training sets. These sets are drawn five times and the performance is averaged out. 

\subsubsection{Evaluation Metrics} We use three metrics for evaluation:
\begin{itemize}
\item Intent Classification Accuracy (ICA): For IC we use accuracy, which is simply the fraction of utterances with correctly predicted intent. 
\item Entity F1 (EF1): For ET, we use standard F1 score calculated by the CoNLL-2003 evaluation script.
\item Sentence Error Rate (SER): We use SER as a metric for joint evaluation of IC and ET to reflect overall erroneous outputs. It is the fraction of utterances with either IC error or at least one ET error (lower is better).
\item For LM evaluation, we use the standard perplexity score.
\end{itemize}

\subsubsection{Significance Test} We use t-test to establish the statistical significance of a method over another (p-value = 0.05).

\begin{figure*}[h]
       \centering
        \begin{subfigure}[b]{0.33\textwidth}
                \centering
                \includegraphics[width=\linewidth]{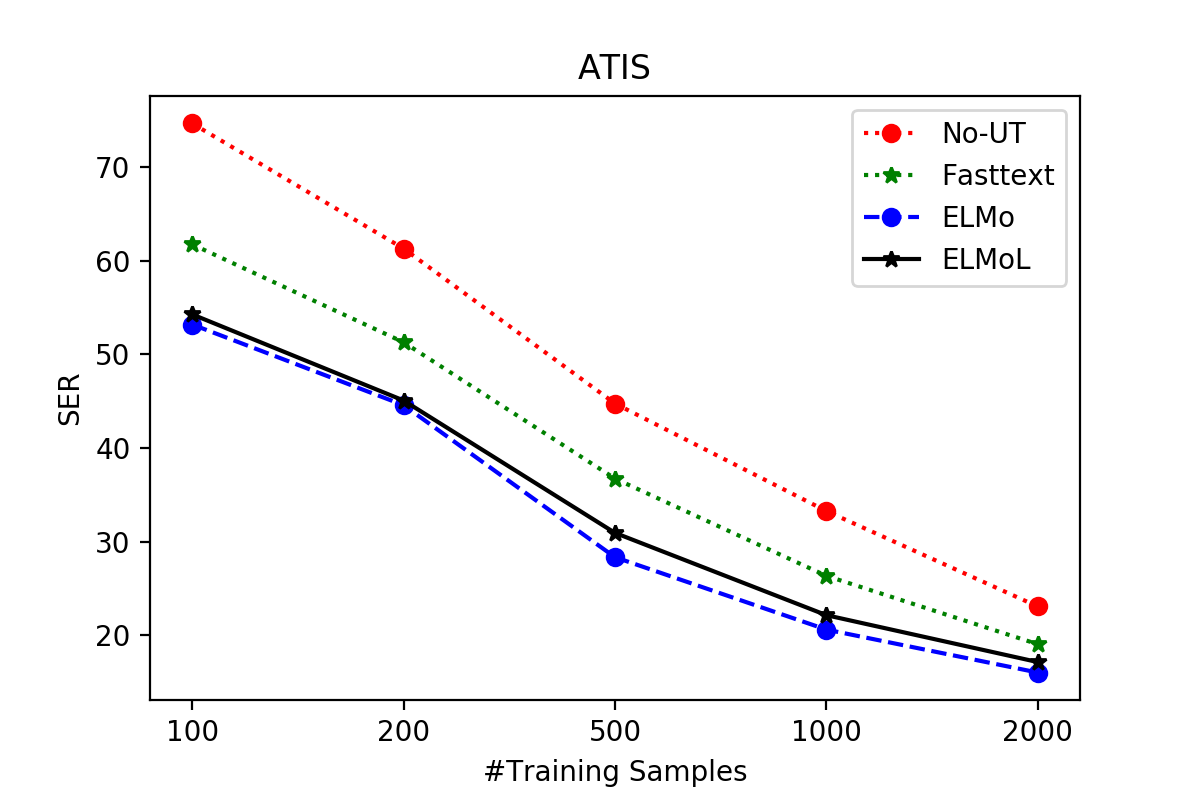}
                \label{fig:atis1}
                \caption{} \label{fig:smalla}
        \end{subfigure}%
       \begin{subfigure}[b]{0.33\textwidth}
                \centering
                \includegraphics[width=\linewidth]{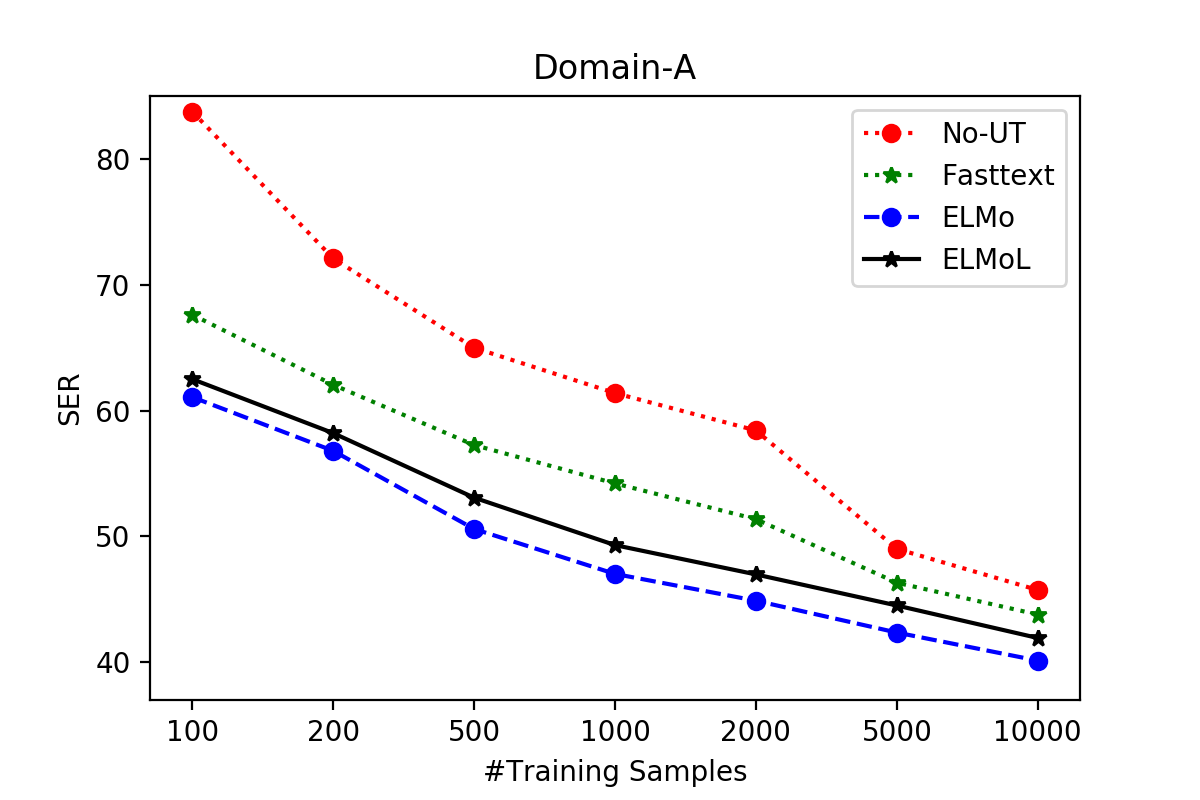}
                \label{fig:recipes1}
                \caption{} \label{fig:smallb}
        \end{subfigure}%
        \begin{subfigure}[b]{0.33\textwidth}
                \centering
                \includegraphics[width=\linewidth]{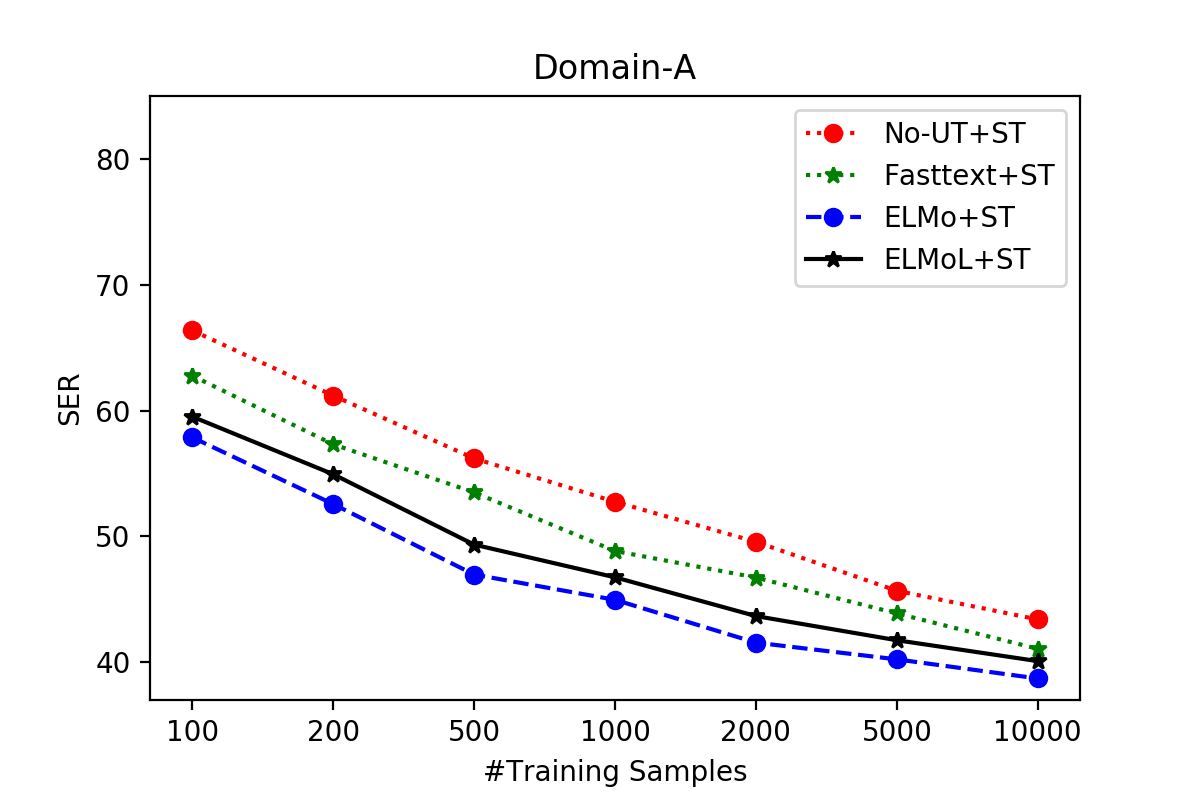}
                \label{fig:recipes4}
                \caption{} \label{fig:smallc}
        \end{subfigure}%
        \\
        \begin{subfigure}[b]{0.33\textwidth}
                \centering
                \includegraphics[width=\linewidth]{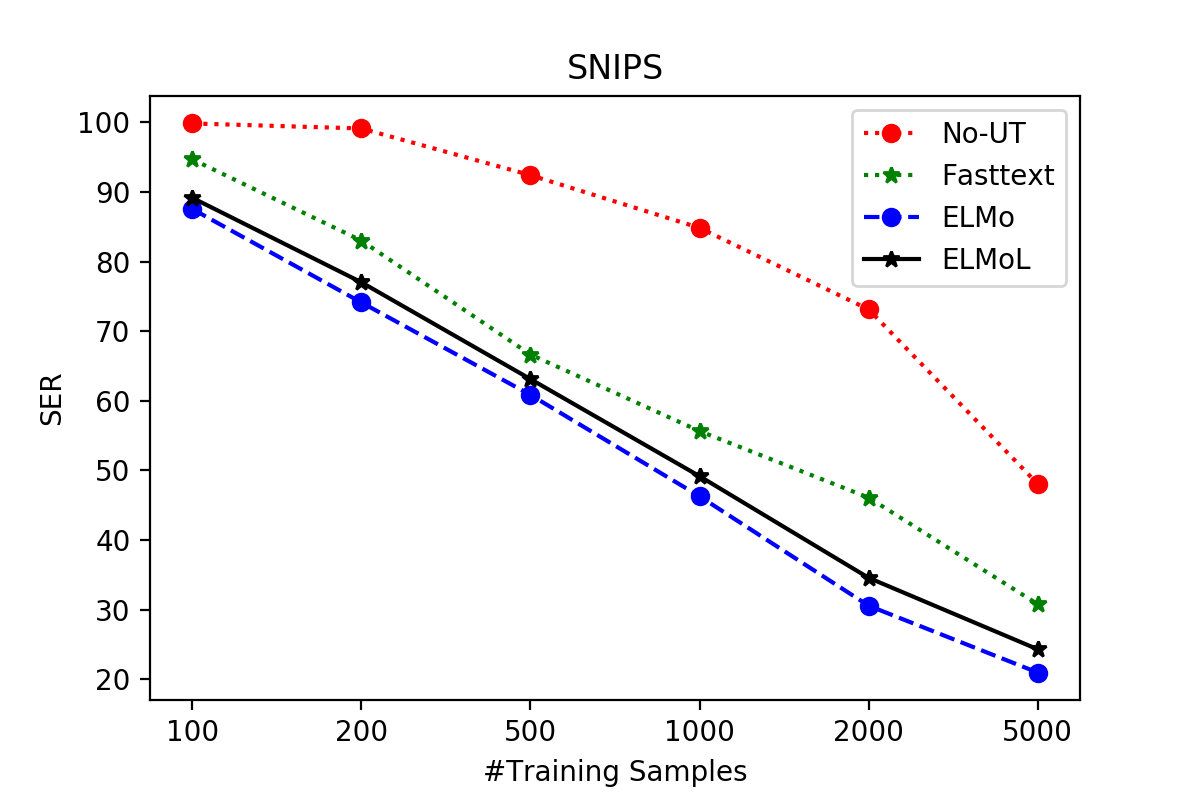}
                \label{fig:snips1}
                \caption{} \label{fig:smalld}
        \end{subfigure}%
        \begin{subfigure}[b]{0.33\textwidth}
                \centering
                \includegraphics[width=\linewidth]{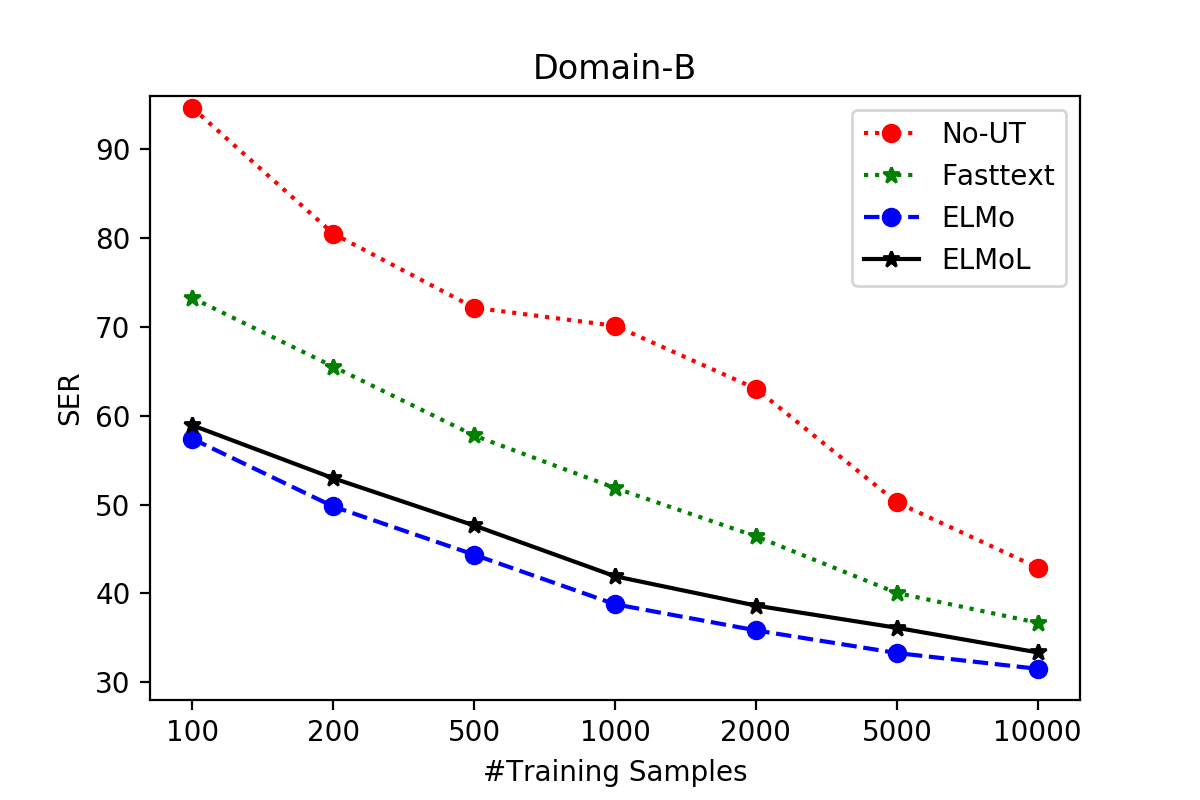}
                \label{fig:localsearch1}
                \caption{} \label{fig:smalle}
        \end{subfigure}%
        \begin{subfigure}[b]{0.33\textwidth}
                \centering
                \includegraphics[width=\linewidth]{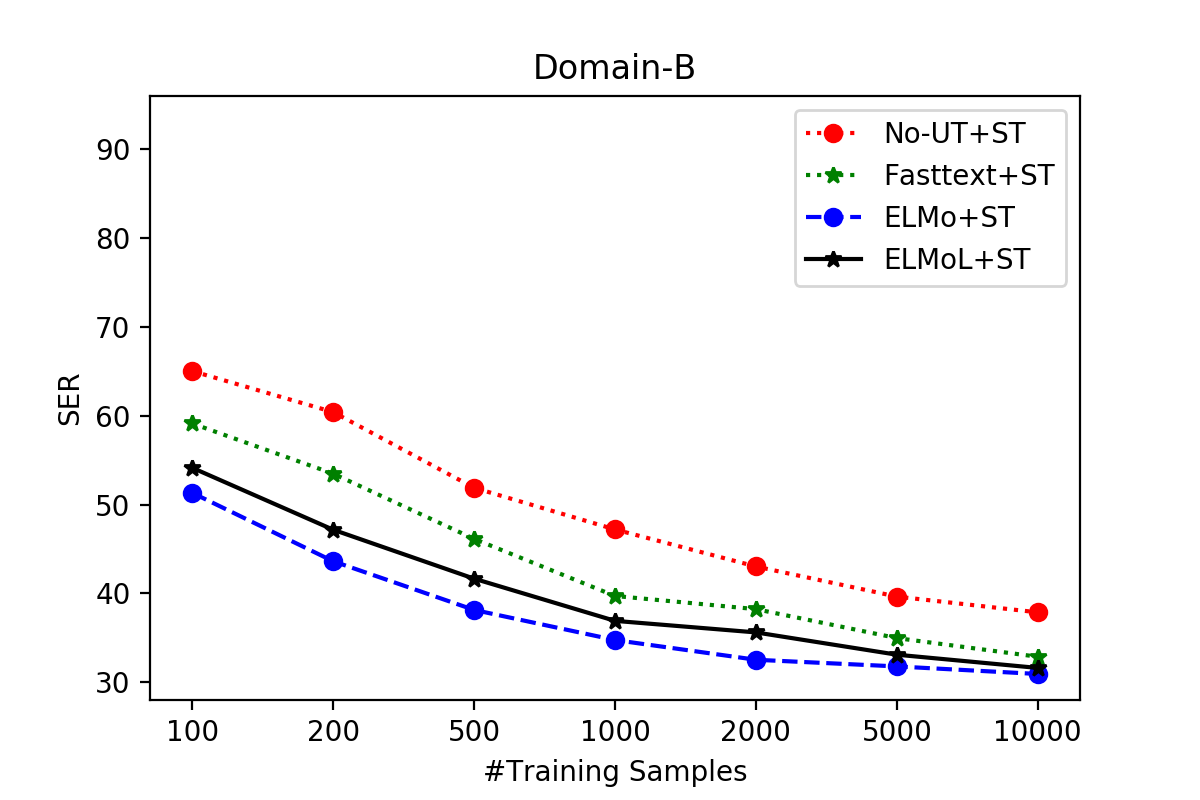}
                \label{fig:localsearch4}
                \caption{} \label{fig:smallf}
        \end{subfigure}%
        \caption{Performance of different methods of UT using varying amounts of training data on ATIS (a), SNIPS (d), Domain-A (b), Domain-B (e). Performance on Domain A and B using UT$+$ST is presented in (c) and (f). The y-scale of (b) and (e) is kept same as (c) and (f) respectively for comparison. Best viewed in color.}
\end{figure*}
% Questions about this plot
% Does this config look good?
% In the rightmost plot should i just plot the 4 *+ST lines and make sure that centre and right have same scale so that they can be compared? 

\subsection{Language Modeling Results}
\label{sec:experiment_lm_results}
Table \ref{table:pplx} shows the perplexity numbers for language models of ELMo and ELMoL trained on different datasets. The upper part of the table shows performance on Internal Datasets, with LMs trained on IUD, 1BWB and their combination and evaluated on held out development sets. Overall, ELMo LM achieves the best performance when first trained on 1BWB and then fine-tuned on either IUD or IMSLU. The ELMoL LM perplexity is higher than ELMo, which is expected because of the much smaller ELMoL LM architecture. However, we find that this difference in perplexity between ELMo and ELMoL does not result in a large SLU performance difference when used in downstream tasks which we discuss in next section.

\begin{table}[h] 
\begin{tabular}{|c|c|c|c|c|}
\hline
\multicolumn{2}{|c|}{Internal Datasets}                                     & \begin{tabular}[c]{@{}c@{}}IUD\\ Holdout\end{tabular} & \begin{tabular}[c]{@{}c@{}}Domain-A\\ Dev\end{tabular} & \begin{tabular}[c]{@{}c@{}}Domain-B\\ Dev\end{tabular} \\ \hline
\multirow{2}{*}{ELMo} & IUD                                                 & 45.2                                                  & 19.4                                                   & 35.8                                                   \\ \cline{2-5} 
                      & \begin{tabular}[c]{@{}c@{}}1BWB+\\ IUD\end{tabular} & 44.9                                                  & 19.1                                                   & 34.9                                                   \\ \hline
ELMoL                 & IUD                                                 & 56.7                                                  & 24.2                                                   & 43.2                                                   \\ \hline
\multicolumn{2}{|c|}{Public Datasets}                                         & \begin{tabular}[c]{@{}c@{}}1MSLU\\ Holdout\end{tabular} & \begin{tabular}[c]{@{}c@{}}ATIS\\ Dev\end{tabular} & \begin{tabular}[c]{@{}c@{}}SNIPS\\ Dev\end{tabular} \\ \hline
\multirow{2}{*}{ELMo} & 1MSLU                                                 & 38.0                                                    & 16.6                                               & 60.7                                                \\ \cline{2-5} 
                      & \begin{tabular}[c]{@{}c@{}}1BWB+\\ 1MSLU\end{tabular} & 31.0                                                    & 15.2                                               & 50.8                                                \\ \hline
ELMoL                 & 1MSLU                                                 & 40.6                                                    & 17.7                                               & 60.2                                                \\ \hline
\end{tabular}
\caption{Perplexity of ELMo and ELMoL language model trained on different datasets.} \label{table:pplx}
\end{table}

\subsection{UT Results for Internal SLU tasks}
\label{sec:results_unsupervised_internal}
The results of Unsupervised Transfer Learning (UT) on internal datasets, Domain A and Domain B are shown in \ref{table:utl_int}. Both ELMo and ELMoL beat the NoUT and FastText baseline significantly on both Domain A and Domain B (p$<$0.05, when comparing ELMo and ELMoL vs NoUT and FastText). ELMo gives a performance improvement of $\sim$ 2 absolute SER points on Domain A and $\sim$ 1.5 absolute SER points on Domain B over Fasttext. With much smaller architecture and 1.6x faster inference, ELMoL does not perform significantly worse compared to ELMo for three out of the four datasets we examined (ATIS, Domains A and B) except SNIPS where ELMo significantly outperforms ELMoL. It is short of ELMo by just $\sim$ 0.3 SER points on Domain A and $\sim$ only 0.5 SER points on Domain B.

In Figures \ref{fig:smallb} and \ref{fig:smalle}, we plot the SER performance as a function of training data size using smaller training sets for Domain A and Domain B. In both plots, we observe the trend that SER performance using 10000 samples with NoUT approximately matches performance using 5000 samples with Fasttext, using 2000 samples with ELMoL and using 1000 samples with ELMo. Hence, unsupervised pre-training reduces the number of labeled target samples required for achieving same level of performance as Fasttext by 5 times using ELMo and 3 times using ELMoL. This enables building accurate models for a new SLU domain faster by reducing labeling time and effort through leveraging UT.

\begin{table}[h]
\scalebox{0.95}{
\begin{tabular}{|c|c|c|c|c|c|c|}
\hline
         & \multicolumn{3}{c|}{Domain-A} & \multicolumn{3}{c|}{Domain-B} \\ \hline
         & ICA       & EF1      & SER      & ICA        & EF1        & SER       \\ \hline
NoUT     & 91.96     & 76.58    & 39.79    & 91.68      & 74.70      & 30.93     \\ \hline
Fasttext & 92.34     & 77.74    & 38.76    & 92.27      & 76.59      & 29.30     \\ \hline
ELMo     & 93.71     & 78.59    & 36.58    & 92.46      & 78.43      & 27.95     \\ \hline
ELMoL    & 93.23     & 78.23    & 36.86    & 92.35      & 78.34      & 28.42     \\ \hline
\end{tabular}}
\caption{Performance of UT methods on internal datasets. Training set sizes are 43K and 100K for domains A and B.}
\label{table:utl_int}
\end{table}

\subsection{UT Results for Benchmark SLU Tasks}
\label{sec:results_unsupervised_external}
Table \ref{table:utl_bm} shows the performance of Unsupervised Transfer on ATIS and SNIPS and Figure \ref{fig:smalla} and \ref{fig:smalld} plot the SER numbers using smaller training sets for these two datasets (low resource simulation). Similar to the trends we observed on internal datasets, we notice that both ELMo and ELMoL significantly outperform the baselines (p$<$0.05, when comparing ELMo and ELMoL vs NoUT and FastText) and ELMoL's performance is only slightly lower than that of ELMo. State-of-the-art (SOTA) for each of the two datasets is presented in the last row. For ATIS, \citet{liu_lane2016} achieve the SOTA by using more complex attention-based models without transfer learning. We are able to get close to their accuracy with simpler models using UT. For SNIPS, we outperform the SOTA performance \citep{snips_url}.

\begin{table}[h]
\centering
\scalebox{0.95}{
\begin{tabular}{|c|c|c|c|c|c|c|}
\hline
         & \multicolumn{3}{c|}{ATIS} & \multicolumn{3}{c|}{SNIPS} \\ \hline
         & ICA  & EF1  & SER   & ICA  & EF1  & SER    \\ \hline
NoUT     & 95.41   & 94.30   & 17.13 & 98.43   & 88.78   & 24.14  \\ \hline
Fasttext & 96.75   & 95.35   & 15.01 & 98.57   & 91.78   & 19.57  \\ \hline
ELMo     & 97.42   & 95.62   & 12.65 & 99.29   & 93.90   & 14.57  \\ \hline
ELMoL    & 97.30   & 95.42   & 13.38 & 98.83   & 93.29   & 15.62  \\ 
\hline
SOTA    &  98.43   & 95.87   & - & -   & 93   & -  \\ 
\hline
\end{tabular}}
\caption{Performance of UT methods on benchmark datasets. SOTA indicates state-of-the-art referring to \citet{liu_lane2016} for ATIS and \citet{snips_url} for SNIPS.} \label{table:utl_bm}
\end{table}

\subsection{ELMoL Training: Ablation Study}
\label{sec:results_ablation}

In Table \ref{table:abl}, we evaluate the gains provided by various training strategies employed for ELMoL to prevent catastrophic forgetting. Vanilla refers to the base ELMoL method, while \textit{guf}, \textit{discr} and \textit{tlr} are training techniques we described in Section \ref{sec:sslmo}. Improvements from each of these techniques are consistent across datasets.

\begin{table}[h]
\centering
\begin{tabular}{|c|c|c|c|c|}
\hline
                                                             & ATIS  & SNIPS & Domain-A & Domain-B \\ \hline
%vanilla                                                      & 14.83 & 18.29 & 38.23    & 29.12    \\ \hline
vanilla                                                  & 14.46 & 17.14 & 37.59    & 28.78    \\ \hline
\textit{guf}                                                      & 13.79 & 16.39 & 37.26    & 28.72    \\ \hline
\begin{tabular}[c]{@{}c@{}} \textit{guf}+\\ \textit{discr+tlr}\end{tabular} & 13.38 & 15.62 & 36.86    & 28.42    \\ \hline
\end{tabular}
\caption{Performance of training techniques in ELMoL. All the above numbers are SER scores.} \label{table:abl}
\end{table}

\subsection{UT+ST Results for SLU}
\label{sec:results_combination}

Table \ref{table:combine_st_ut} shows the performance when combining each of the UT methods presented earlier along with supervised transfer (ST). Comparing with the performance of Table \ref{table:utl_int}, we see a consistent improvement of all UT methods when combining with ST. Specifically, ELMo+ST achieves SER 35.86 and 27.46 compared to 36.58 and 27.95 for ELMo for Domains A and B respectively. Similarly, ELMoL+ST achieves SER 36.38 and 28.11 compared to 36.86 and 28.42 for ELMoL for Domains A and B respectively. This indicates that gains from UT and ST are additive, and transferring knowledge from all available data, both labeled and unlabeled, is beneficial for the final SLU accuracy.

\begin{table}[h]
\scalebox{0.9}{
\begin{tabular}{|c|c|c|c|c|c|c|}
\hline
            & \multicolumn{3}{c|}{Domain-A} & \multicolumn{3}{c|}{Domain-B} \\ \hline
            & ICA       & EF1      & SER      & ICA        & EF1        & SER       \\ \hline
NoUT+ST     & 92.80     & 76.85    & 38.97    & 91.57      & 75.53      & 30.32     \\ \hline
Fasttext+ST & 93.15     & 77.27    & 37.58    & 92.36      & 76.73      & 28.95     \\ \hline
ELMo+ST     & 93.79     & 78.76    & 35.86    & 92.88      & 78.49      & 27.46     \\ \hline
ELMoL+ST    & 93.53     & 78.56    & 36.38    & 92.84      & 77.03      & 28.11     \\ \hline
\end{tabular}}
\caption{Performance when ST is combined with UT.} \label{table:combine_st_ut}
\end{table}

\subsection{ELMo vs ELMoL: Performance Comparison}
The proposed ELMoL model is faster in terms of both training time and inference time, and smaller in total model size compared to standard ELMo. Assuming ELMoL takes 1 unit of time, ELMo would take 1.8x more time to train and 1.6x more time during inference. Fast training and inference times are critical in deployed SLU systems as they enable rapid domain development and timely response to the user’s request during runtime. Also, ELMoL requires 4x lesser memory than ELMo (37 million parameters in ELMoL compared to 140 million in model with ELMo), which is advantageous for SLU systems deployed in embedded devices or other resource constrained systems.

\section{Conclusions and Future Work}
\label{sec:conclusions_future}

We describe unsupervised transfer (UT) learning techniques for leveraging unlabeled text data for SLU tasks. Specifically, we showed that the recently proposed ELMo embeddings \cite{peters_elmo_2018} improve IC and ET accuracy in multi-task setting. We also proposed a light-weight alternative to ELMo, called ELMo-Light (ELMoL). We showed that ELMoL performance is comparable to ELMo, and it is faster at runtime and better suited for practical SLU systems. Our UT techniques achieved large gains over using only labeled in-domain data for training, with gains being more pronounced for low resource settings. Finally, we also showed that gains from ST and UT are additive. In future, we plan to apply the transfer techniques across different languages. We would also like to experiment with alternative architectures such as transformer and adversarial networks. 

\bibliography{unsupervised}

\begin{thebibliography}{}

\bibitem[\protect\citeauthoryear{Bojanowski \bgroup et al\mbox.\egroup
  }{2017}]{bojanowski2017enriching}
Bojanowski, P.; Grave, E.; Joulin, A.; and Mikolov, T.
\newblock 2017.
\newblock Enriching word vectors with subword information.
\newblock {\em Transactions of the Association of Computational Linguistics
  (TACL)} 5(1):135--146.

\bibitem[\protect\citeauthoryear{Cer \bgroup et al\mbox.\egroup
  }{2018}]{Cer2018universal}
Cer, D.; Yang, Y.; Kong, S.-Y.; Hua, N.; Limtiaco, N.; John, R.~S.; Constant,
  N.; Guajardo-Cespedes, M.; Yuan, S.; Tar, C.; Sung, Y.-H.; Strope, B.; and
  Kurzweil, R.
\newblock 2018.
\newblock Universal sentence encoder.
\newblock {\em arXiv preprint arXiv:1803.11175}.

\bibitem[\protect\citeauthoryear{Chelba \bgroup et al\mbox.\egroup
  }{2014}]{chelba2014_1BWB}
Chelba, C.; Mikolov, T.; Schuster, M.; Ge, Q.; Brants, T.; Koehn, P.; and
  Robinson, T.
\newblock 2014.
\newblock One billion word benchmark for measuring progress in statistical
  language modeling.
\newblock In {\em Interspeech}.

\bibitem[\protect\citeauthoryear{Chiu and Nichols}{2016}]{chiu2016named}
Chiu, J.~P., and Nichols, E.
\newblock 2016.
\newblock Named entity recognition with bidirectional lstm-cnns.
\newblock {\em Transactions of the Association for Computational Linguistics
  (TACL)} 4:357--370.

\bibitem[\protect\citeauthoryear{Collobert and
  Weston}{2008}]{collobert_weston_08}
Collobert, R., and Weston, J.
\newblock 2008.
\newblock A unified architecture for natural language processing: Deep neural
  networks with multitask learning.
\newblock In {\em International Conference of Machine Learning (ICML)}.

\bibitem[\protect\citeauthoryear{Coucke \bgroup et al\mbox.\egroup
  }{2017}]{snips_url}
Coucke, A.; Ball, A.; Delpuech, C.; Doumouro, C.; Raybaud, S.; Gisselbrecht,
  T.; and Dureau, J.
\newblock 2017.
\newblock Benchmarking natural language understanding systems: Google,
  facebook, microsoft, amazon and snips.

\bibitem[\protect\citeauthoryear{Dai and Le}{2015}]{dai_le_2015}
Dai, A.~M., and Le, Q.~V.
\newblock 2015.
\newblock Semi-supervised sequence learning.
\newblock In {\em Neural Information Processing Systems (NIPS)}.

\bibitem[\protect\citeauthoryear{Donahue \bgroup et al\mbox.\egroup
  }{2014}]{donahue2014decaf}
Donahue, J.; Jia, Y.; Vinyals, O.; Hoffman, J.; Zhang, N.; Tzeng, E.; and
  Darrell, T.
\newblock 2014.
\newblock Decaf: A deep convolutional activation feature for generic visual
  recognition.
\newblock In {\em International Conference on Machine Learning (ICML)}.

\bibitem[\protect\citeauthoryear{Goyal, Metallinou, and
  Matsoukas}{2018}]{goyal_tl_2018}
Goyal, A.; Metallinou, A.; and Matsoukas, S.
\newblock 2018.
\newblock Fast and scalable expansion of natural language understanding
  functionality for intelligent agents.
\newblock In {\em North American Chapter of the Association for Computational
  Linguistics: Human Language Technologies (NAACL-HLT)}.

\bibitem[\protect\citeauthoryear{Hemphill, Godfrey, and
  Doddington}{1990}]{hemphill1990atis}
Hemphill, C.~T.; Godfrey, J.~J.; and Doddington, G.~R.
\newblock 1990.
\newblock The atis spoken language systems pilot corpus.
\newblock In {\em Speech and Natural Language Workshop}.

\bibitem[\protect\citeauthoryear{Henderson, Thomson, and
  Williams}{2014}]{henderson2014second}
Henderson, M.; Thomson, B.; and Williams, J.~D.
\newblock 2014.
\newblock The second dialog state tracking challenge.
\newblock In {\em Special Interest Group on Discourse and Dialogue (SIGDIAL)}.

\bibitem[\protect\citeauthoryear{Howard and Ruder}{2018}]{howard_ruder_2018}
Howard, J., and Ruder, S.
\newblock 2018.
\newblock Universal language model fine-tuning for text classification.
\newblock In {\em Association for Computational Linguistics(ACL)}.

\bibitem[\protect\citeauthoryear{Jozefowicz \bgroup et al\mbox.\egroup
  }{2016}]{jozefowicz2016exploring}
Jozefowicz, R.; Vinyals, O.; Schuster, M.; Shazeer, N.; and Wu, Y.
\newblock 2016.
\newblock Exploring the limits of language modeling.
\newblock {\em arXiv preprint arXiv:1602.02410}.

\bibitem[\protect\citeauthoryear{Kim \bgroup et al\mbox.\egroup
  }{2017}]{kim_emnlp_2017}
Kim, J.-K.; Kim, Y.-B.; Sarikaya, R.; and Fosler-Lussier, E.
\newblock 2017.
\newblock Cross-lingual transfer learning for pos tagging without cross-lingual
  resources.
\newblock In {\em Empirical Methods in Natural Language Processing (EMNLP)}.

\bibitem[\protect\citeauthoryear{Kinmga and Ba}{2015}]{kingma2015method}
Kinmga, D., and Ba, J.
\newblock 2015.
\newblock A method for stochastic optimization.
\newblock In {\em International Conference on Learning Representations (ICLR)}.

\bibitem[\protect\citeauthoryear{Kiros \bgroup et al\mbox.\egroup
  }{2015}]{kiros_skipthought_2015}
Kiros, R.; Zhu, Y.; Salakhutdinov, R.~R.; Zemel, R.; Urtasun, R.; Torralba, A.;
  and Fidler, S.
\newblock 2015.
\newblock Skip-thought vectors.
\newblock In {\em Neural Information Processing Systems (NIPS)}.

\bibitem[\protect\citeauthoryear{Krizhevsky, Sutskever, and
  Hinton}{2012}]{krizhevsky2012imagenet}
Krizhevsky, A.; Sutskever, I.; and Hinton, G.~E.
\newblock 2012.
\newblock Imagenet classification with deep convolutional neural networks.
\newblock In {\em Neural Information Processing Systems (NIPS)}.

\bibitem[\protect\citeauthoryear{Liu and Lane}{2016}]{liu_lane2016}
Liu, B., and Lane, I.
\newblock 2016.
\newblock Attention-based recurrent neural network models for joint intent
  detection and slot filling.
\newblock In {\em Interspeech}.

\bibitem[\protect\citeauthoryear{Liu \bgroup et al\mbox.\egroup
  }{2015}]{liu_multi_timescale}
Liu, P.; Qiu, X.; Chen, X.; Wu, S.; and Huang., X.
\newblock 2015.
\newblock Multi-timescale long short-term memory neural network for modelling
  sentences and documents.
\newblock In {\em Empirical Methods for Natural Language Processing (EMNLP)}.

\bibitem[\protect\citeauthoryear{Liu \bgroup et al\mbox.\egroup
  }{2018}]{liu_lm_2018}
Liu, L.; Shang, J.; Xu, F.; Ren, X.; Gui, H.; Peng, J.; and Han, J.
\newblock 2018.
\newblock Empower sequence labeling with task-aware neural language model.
\newblock In {\em Association for the Advancement of Artificial Intelligence
  (AAAI)}.

\bibitem[\protect\citeauthoryear{Mikolov \bgroup et al\mbox.\egroup
  }{2013}]{mikolov_w2v_2013}
Mikolov, T.; Sutskever, I.; Chen, K.; Corrado, G.; and Dean, J.
\newblock 2013.
\newblock Distributed representations of words and phrases and their
  compositionality.
\newblock In {\em Neural Information Processing Systems (NIPS)}.

\bibitem[\protect\citeauthoryear{Peters \bgroup et al\mbox.\egroup
  }{2018}]{peters_elmo_2018}
Peters, M.~E.; Neumann, M.; Iyyer, M.; Gardner, M.; Clark, C.; Lee, K.; and
  Zettlemoyer, L.
\newblock 2018.
\newblock Deep contextualized word representations.
\newblock In {\em North American Chapter of the Association for Computational
  Linguistics: Human Language Technologies (NAACL-HLT)}.

\bibitem[\protect\citeauthoryear{Radford \bgroup et al\mbox.\egroup
  }{2018}]{Radford_2018}
Radford, A.; Narasimhan, K.; Salimans, T.; and Sutskever, I.
\newblock 2018.
\newblock Improving language understanding by generative pre-training.
\newblock {\em Preprint}.

\bibitem[\protect\citeauthoryear{Sharif~Razavian \bgroup et al\mbox.\egroup
  }{2014}]{sharif2014cnn}
Sharif~Razavian, A.; Azizpour, H.; Sullivan, J.; and Carlsson, S.
\newblock 2014.
\newblock Cnn features off-the-shelf: an astounding baseline for recognition.
\newblock In {\em Computer Vision and Pattern Recognition (CVPR) Workshop}.

\bibitem[\protect\citeauthoryear{Socher \bgroup et al\mbox.\egroup
  }{2013}]{Socher_recursive_2013}
Socher, R.; Perelygin, A.; Wu, J.~Y.; Chuang, J.; Manning, C.~D.; Ng, A.~Y.;
  and Potts, C.
\newblock 2013.
\newblock Recursive deep models for semantic compositionality over a sentiment
  treebank.
\newblock In {\em Empirical Methods for Natural Language Processing (EMNLP)}.

\end{thebibliography}
\bibliographystyle{aaai}

\end{document}